%% file: root.tex
\documentclass[letterpaper, 10 pt, conference]{ieeeconf}  %

\IEEEoverridecommandlockouts                              %

\usepackage{lipsum}

\usepackage{amsfonts,amssymb,amsmath,bm}

\usepackage{graphicx}
\usepackage{siunitx}
\usepackage{glossaries}

\usepackage{color}
\usepackage{ifthen}
\usepackage{flushend}
\newboolean{authnotes}
\usepackage{subfigure}
\usepackage{tabularx}
\usepackage{multirow}
\usepackage{multicol}
\usepackage{makecell}
\usepackage{booktabs}
\usepackage{comment}
\usepackage{hyperref}

\makeatletter
\let\NAT@parse\undefined
\makeatother

\setboolean{authnotes}{true}

\ifthenelse{\boolean{authnotes}}
{
\newcommand{\todo}[1]{{{\bf\color{magenta} TODO: #1}}}

}
{
\newcommand{\todo}[1]{}
\newcommand{\nf}[1]{}
\newcommand{\db}[1]{}
\newcommand{\st}[1]{}
}

\newcommand*{\Scale}[2][4]{\scalebox{#1}{$#2$}}%

\let\ACMmaketitle=\maketitle
\renewcommand{\maketitle}{\begingroup\let\footnote=\thanks \ACMmaketitle\endgroup}

\makeatletter
\newcommand*\titleheader[1]{\begingroup\gdef\@titleheader{#1}\let\footnote=\thanks\endgroup}
\AtBeginDocument{%
  \let\st@red@title\@title
  \def\@title{%
  \begin{flushleft}
    \vspace{-2.0em}
    \bgroup\normalfont\small\@titleheader\par\egroup
    \vspace{-18pt}\par\noindent\rule{\textwidth}{0.1pt}
    \end{flushleft}
    \vskip0.5em\st@red@title
        }

}

\makeatother

\input{misc/math_commands}

\title{\LARGE \bf
Graph-based Reinforcement Learning meets Mixed Integer Programs: An application to 3D robot assembly discovery
}

\author{Niklas Funk, Svenja Menzenbach, Georgia Chalvatzaki, and Jan Peters%
\thanks{*This work is supported by the AICO grant by the Nexplore/Hochtief Collaboration with TU Darmstadt, and the Emmy Noether DFG Programme (No. 448644653). Calculations for this research were conducted on the Lichtenberg high performance computer of the TU Darmstadt.}%
\thanks{ Department of Computer Science,
        Technical University of Darmstadt,
        {\tt\small \{niklas,georgia,jan\}@robot-learning.de}}%
\thanks{This work has been submitted to the IEEE for possible publication. Copyright may be transferred without notice, after which this version may no longer be accessible.}
}

\titleheader{\textit{Preprint}}

\begin{document}

\newacronym{milp}{MILP}{mixed-integer linear program}
\newacronym{gnn}{GNN}{graph neural network}
\newacronym{mdp}{MDP}{Markov Decision Process}
\newacronym{mcts}{MCTS}{Monte Carlo Tree Search}
\newacronym{tamp}{TAMP}{Task and Motion Planning}
\newacronym{rl}{RL}{Reinforcement Learning}
\newacronym{nn}{NN}{neural network}
\newacronym{ik}{IK}{inverse kinematics}
\newacronym{gamp}{GAMP}{grasp and motion planning}
\newacronym{rad}{RAD}{\textit{robotic assembly discovery}}

\maketitle
\thispagestyle{empty}
\pagestyle{empty}

\begin{abstract}
\textit{Robot assembly discovery (RAD)} is a challenging problem that lives at the intersection of resource allocation and motion planning.
The goal is to combine a predefined set of objects to form something new while considering task execution with the robot-in-the-loop.
In this work, we tackle the problem of building arbitrary, predefined target structures entirely from scratch using a set of Tetris-like building blocks and a robotic manipulator. 
Our novel hierarchical approach aims at efficiently decomposing the overall task into three feasible levels that benefit mutually from each other.
On the high level, we run a classical mixed-integer program for global optimization of block-type selection and the blocks' final poses to recreate the desired shape. Its output is then exploited to efficiently guide the exploration of an underlying reinforcement learning (RL) policy.
This RL policy draws its generalization properties from a flexible graph-based representation that is learned through Q-learning and can be refined with search. Moreover, it accounts for the necessary conditions of structural stability and robotic feasibility that cannot be effectively reflected in the previous layer.
Lastly, a grasp and motion planner transforms the desired assembly commands into robot joint movements.
We demonstrate our proposed method's performance on a set of competitive simulated RAD environments, showcase real-world transfer, and report performance and robustness gains compared to an unstructured end-to-end approach.  

\end{abstract}

\section{INTRODUCTION}

A common desire amongst many industry sectors is to increase resource efficiency. The construction industry is a key sector that could significantly reduce its environmental impact by re-using existing material more efficiently, moving towards the ideas of circular economy \cite{durmisevic2019circular}. 
There is a fundamental need for combining intelligent algorithms for reasoning on how existing material can be recombined to form something new, with autonomous execution \cite{tibbits2017autonomous}.

\begin{figure}
    \centering
    \includegraphics[width=1.0\columnwidth]{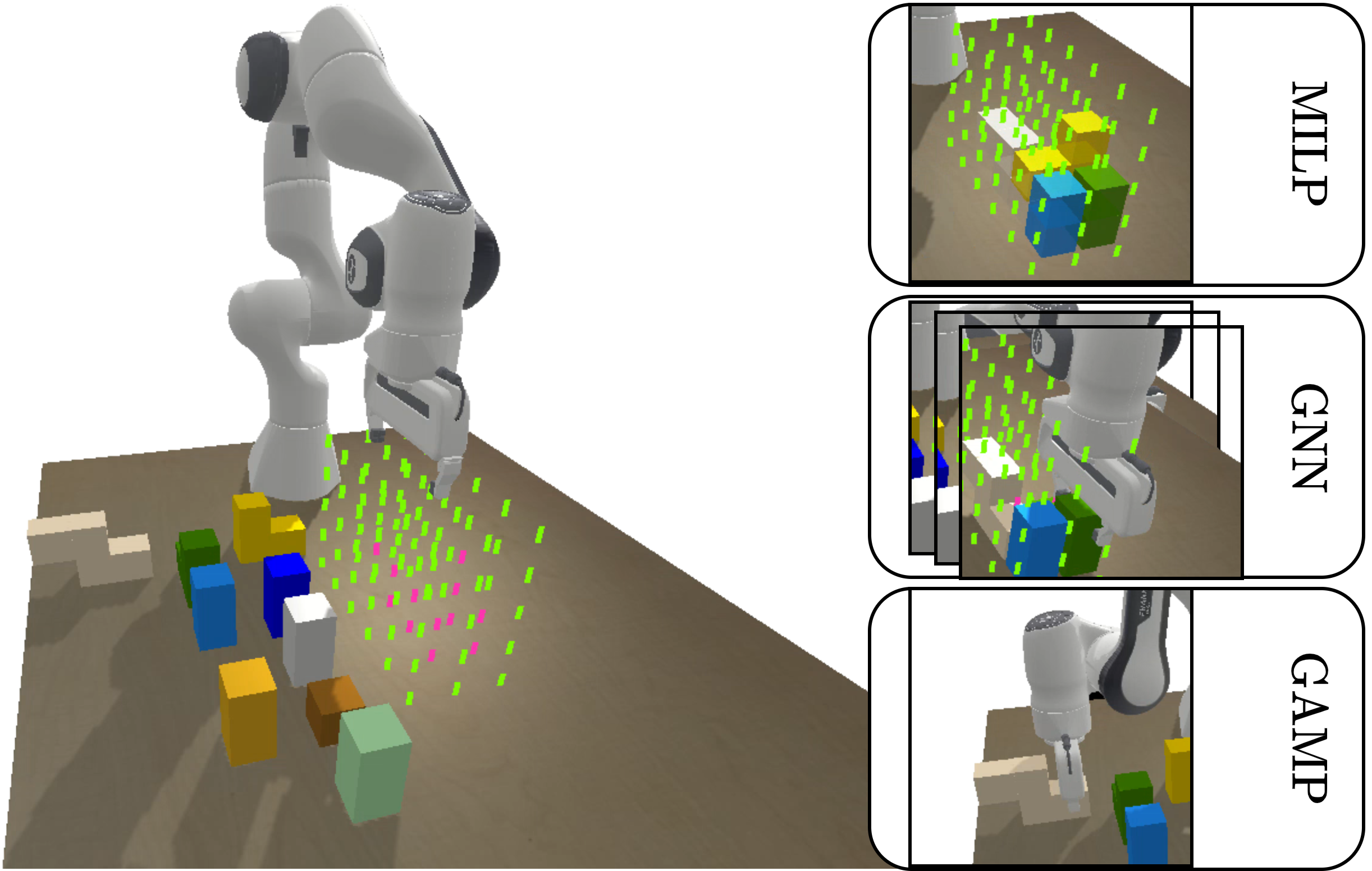}
    \caption{\small Simulated \gls{rad} environment (left) and all three components of our proposed hierarchical approach (right). On the highest level, we solve an MILP to determine the building blocks' poses to optimally fill the voxelized, desired target shape (visualized in pink/green). Next follows a learned GNN policy determining which block to move next based on the scene information and MILP solution. Lastly, we run a GAMP to determine how to grasp the chosen block and realize the robot movement on the joint level.}
    \label{fig:overview_alt}
    \vspace{-0.7cm}
\end{figure}

Herein, we are concerned with the problem of autonomous \gls{rad}, where a robotic agent should reason about abstract 3D target shapes that need to be fulfilled given a set of available building blocks (cf. Fig. \ref{fig:overview_alt}). Unlike other assembly problems with known instructions, in \gls{rad}, the agent does neither have any prior information about which blocks to use and their final poses, 
nor about the execution sequence.
Contrarily, the \gls{rad} agent should \textit{discover} the possible ways of combining the building blocks, find appropriate action sequences, and put them into practice.
\gls{rad} can thus be structured into two difficulty levels.
On the high level, a goal-defined resource allocation problem has to be solved, which is typically NP-complete for discrete resources, and can be viewed as a real-world version of the Knapsack Problem \cite{salkin1975knapsack}.
The low level requires solving a constrained motion planning problem, considering kinematic feasibility and structural stability.
\\
One way of approaching this problem are end-to-end approaches that directly map from problem definition to low level actions \cite{hamrick2019combining, funk22a}.
Such approaches are typically straightforward to design, and draw their generalization properties from learned graph-based representations. 
Yet, they often require extensive training due to the huge combinatorial action space, %
and are typically hard to debug and interpret. 
On the other end of the spectrum are \gls{tamp} approaches \cite{toussaint2015logic, kaelbling2010hierarchical}, which naturally represent the problem's hierarchical nature and necessitate full prior knowledge of geometry and kinematics. Yet, they are usually unsuitable for real-time reactive control, as the full joint optimization suffers from combinatorics and non-convex constraints. 

We propose a novel hierarchical and hybrid method for 3D \gls{rad} that addresses both, resource allocation and motion planning. 
Namely, on the high level, a model-based \gls{milp}, handling the process of block-type selection and optimizing the blocks' final poses for optimally resembling the desired target shape, is solved. 
The \gls{milp}'s solution is then used as a \textit{guiding exploration} signal in a graph-based \gls{rl} framework. 
We define a \gls{gnn} for capturing the geometric, structural, and physical relationships between building blocks, robot, and target shape, thereby incorporating all effects that have not been modelled on the higher level.  
The \gls{gnn} is trained through model-free Q-learning and can, therefore, efficiently reason about the action sequencing, besides enabling the integration with tree search for improved long-term decisions \cite{silver2017mastering}.
To put the previous reasoning into practice, at the lowest level, we rely on a simple \gls{gamp} method \cite{vahrenkamp2010integrated} jointly optimizing grasp pose selection and end-pose motion generation.

To summarize, our proposed approach
    \textbf{(i)} benefits from the combination of global structural reasoning together with local, sequential decision-making,
    \textbf{(ii)} deals efficiently with the huge action space by skipping the complexity of determining the assembly sequencing on the high level, though inducing a strong \textit{inductive bias} for the \gls{rl} problem while exploiting \gls{gamp} at the lowest level,
    \textbf{(iii)} allows transfer and generalization to instances with different target shapes and types/numbers of blocks, as all levels are invariant to problem size, and
    \textbf{(iv)} provides the flexibility of adding search to further increase reliability and robustness. 
We present an empirical evaluation of our proposed approach in a set of competitive simulated \gls{rad} tasks and demonstrate real-world transfer. The results show superior performance against both empirical and learned baselines, thereby underlining its effectiveness. 

\section{RELATED WORKS}

Due to their practical relevance, assembly and resource allocation tasks are studied amongst a wide area of communities. 
Researchers have proposed methods based on \gls{tamp} \cite{toussaint2015logic, kaelbling2010hierarchical} to solve the problem of motion generation for long-horizon tasks \cite{fox2003pddl2}.
The tasks involve the combination of multiple manoeuvres, such as handovers, tool use, or even multi-robot coordination \cite{ren2021extended, hartmann2021long}. Yet, \gls{tamp} suffers from a combinatorial barrier in the search space and high computational costs as 
\gls{tamp} attempts the full optimization of a hybrid problem, in which the high level variables influence the lower level through constraints.
Thus, \gls{tamp} approaches rely on full prior knowledge about geometry, kinematics, and desired goal state. 
To reduce the combinatorial barrier, recently, methods combining learned heuristics with classical optimization were proposed \cite{garrett2016learning, 21driessIJRR, NoseworthyBrandMosesRSS21,silver2021planning}.
In contrast to \gls{tamp} problems, \gls{rad} comes at the additional complexity of not only having to decide on the sequencing of the placement actions and generating feasible motions, but also requires the optimization for the part placement poses. %

Another line of research exclusively investigates the problem of optimal part placement \cite{fasano2015modeling, fasano2014solving, junqueira2012optimization} due to its practical relevance in logistics.
A common theme along these works is the main focus on placing rectilinear objects into a convex domain for which they present \gls{milp} formulations.
There are two types of formulations: position-based ones in which all objects can be placed at continuous positions, and grid-based ones based that voxelize the packing volume.
The latter formulations are usually more practical as their relaxed solutions are tighter, simplifying the challenge of finding an integer feasible solution \cite{fasano2015modeling}.  
The authors of \cite{fasano2015modeling} introduce formulations for both cases and provide a discussion on how irregular items can be approximated with Tetris-like shapes, while \cite{junqueira2012optimization} presents extensions for incorporating additional constraints. %
Recently, \cite{wang2020robot} investigated the problem of bin-packing irregular-shaped objects arriving in a nondeterministic order.
While they present effective heuristic strategies, taking time and sequencing into account, they do not consider the physical action execution with a robot. 

Reasoning about how to combine elements to resemble a given structure is also a crucial challenge amongst the machine learning community \cite{battaglia2018relational}.
Variants of this problem have been explored in the context of \gls{rl}, e.g. in \cite{janner2018reasoning, bapst19a, hamrick2019combining}. In particular, \glspl{gnn} in \gls{rl} frameworks are very promising. Due to their ability to learn relational encodings \cite{vaswani2017attention,velivckovic2017graph} and invariant representations, they can overcome combinatorial barriers \cite{kool2018attention}, and be combined with search for improved generalization and robustness \cite{funk22a,hamrick2019combining,silver2017mastering}. Indeed, in our prior work \cite{funk22a}, we proposed a novel multi-head attention \gls{gnn}, that when trained with Q-learning and combined with \gls{mcts}, can effectively learn to solve \gls{rad} instances in an end-to-end fashion. Yet, the action-space combinatorics induce a very hard exploration problem and thus limit end-to-end \gls{rad}.
Apart from our own work, the previous methods do not consider 3D scenarios and robotic part placement.

Contrarily, \cite{stevvsic2020learning} proposes a method for local assembly from camera images through template prediction.
Similarly, \cite{li2020towards} focuses on building block towers of variable heights through structured representations and model-free \gls{rl}. The authors of \cite{lin2021efficient} also address block-stacking but use learning from demonstrations to train two \glspl{gnn} to split the tasks of selecting the next object and the respective pre-defined goal location. 
\cite{zakka2020form2fit} learn an end-to-end policy from human demonstrations for executing sequential picking and placing given shape correspondences.
A method for next-best pose estimation for stone stacking is presented in \cite{7989272}. 
While these works all consider robotic execution, they do not consider the additional challenge of placing versatile blocks to achieve a desired global goal configuration, which is only specified on an abstract level and does not reveal how the individual parts need to be arranged.

To the best of our knowledge, only our previous work \cite{funk22a} addresses \gls{rad} from a robot learning perspective. 
Yet, herein, we explore a different direction through proposing a novel structured, hierarchical approach by fusing global \gls{milp} optimization with learning local graph-based \gls{rl} assembly policies, that combined with low level \gls{gamp} can reliably deal with complex \gls{rad} instances.

\section{PROBLEM DEFINITION}
\label{sec:problem_def}

\begin{figure}
    \centering
    \includegraphics[width=0.7\columnwidth]{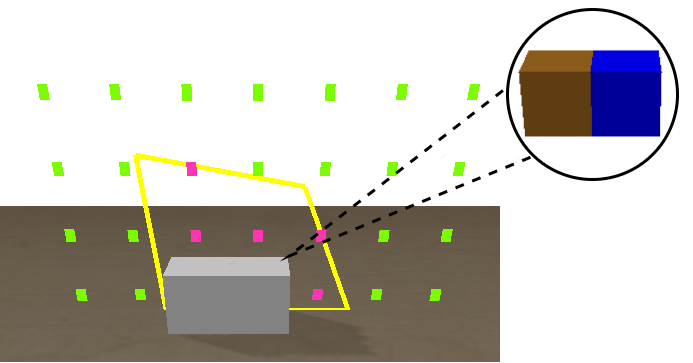}
    \vspace{-0.15cm}
    \caption{Simplified 2D \gls{rad} scene with one placed block consisting of two primitive units (underlined by additionally visualizing them in brown and blue in the top right). Those two primitive units form the set of placed elements $\mathcal{S}_P$. In yellow, we showcase the area that is to be filled, which is discretized through a grid. The grid cells are visualized through their centre points. The pink points correspond to the set of target grid-cells ($\mathcal{T}_F$), i.e. the set of cells that should still be filled as they are part of the desired shape, while the green points represent the so-called non-target grid-cells ($\mathcal{T}_E$) that should ideally remain unoccupied as they are not part of the target shape and as another goal is to also avoid unnecessary placements.}
    \label{fig:grid}
    \vspace{-0.5cm}
\end{figure}

We formulate the problem of having to combine a set of rectlinear, Tetris-like blocks into a desired target shape (cf. Fig. \ref{fig:overview_alt}) as a \gls{mdp} ($\mathcal{S},\mathcal{A},\mathcal{P},r,\gamma$) \cite{puterman2014markov} with state and action space, $\mathcal{S},\mathcal{A}$, transition probabilities $p$, reward function $r$, and discount factor $\gamma$.
To capture the assembly scene's current configuration, the state $s$ is given by the combination of four sets, $s=(\mathcal{S}_U,\mathcal{S}_P,\mathcal{T}_F,\mathcal{T}_E)$, with $|\mathcal{S}_U|{=}N_U, |\mathcal{S}_P|{=}N_P, |\mathcal{T}_F|{=}N_F, |\mathcal{T}_E|{=}N_E$, where
the set $\mathcal{S}_U$ encodes the unplaced primitive units that are still available for construction, $\mathcal{S}_P$ the primitive units that have already been used, $\mathcal{T}_F$ and $\mathcal{T}_E$ contain the so-called target grid-cells and non-target grid-cells, respectively. As shown in Fig. \ref{fig:grid}, we voxelize the building area around the desired target shape and, thus, end up with these grid-cells, which are parameterized through their respective 3D center coordinate $\vx \in \mathbb{R}^3$, i.e. $\mathcal{T}_F{=}\{\vx_i ,i \in N_F\}$, $\mathcal{T}_E{=}\{\vx_i ,i \in N_E\}$ (visualized in pink and green).
While the target grid-cells (pink) are part of the target shape and should ideally all be filled during \gls{rad}, the non-target grid-cells (green) should remain unoccupied.
By projecting all grid-cells centre coordinate $\vx_i$ into the yellow target shape, we decide whether it should be occupied or remain unoccupied during \gls{rad}.  

In this work, we assume that all building blocks are a combination of primitive units. More specifically, we consider that there is only one type of primitive unit: a unit cube $u_c=1^3$. Thus, all the blocks in the scene are a combination of primitive units (cf. Fig. \ref{fig:grid}), i.e., block $i$ is defined by the union of $N_{b_i}$ primitive units, $b_i=\bigcup_{j=1}^{N_{b_i}}u_c$.
Representing blocks as concatenations of primitives allows for a universal interface with graph-based representations, as any Tetris-like block can be easily represented.
Simply put, each primitive unit induces a node in the graph, and the connectivity information encodes whether or not multiple primitive units form a lager block (cf. two leftmost frames in Fig. \ref{fig:gnn}).
This choice also allows us to describe the placed and unplaced blocks through the primitive units' 3D positions $\vx_k$, and connectivity information $y_k=[y_{k,1},...,y_{k,N_U}]$, i.e., $\mathcal{S}_U{=}\{(\vx_k,y_k),k \in N_U\}$. If primitive unit k is connected with primitive unit 1 to form a larger block $y_{k,1}$ equals $1$, otherwise $y_{k,1}{=}0$. We follow the same procedure for the set of already placed elements $\mathcal{S}_P$. 

For placing blocks in the scene, we use a discrete, time-varying action space. In particular, every primitive unit which is at the moment unplaced, can be placed w.r.t. all available grid-cells. As more complicated blocks might also require rotations, we augment all placement actions with four rotational actions, i.e. rotating the block by $0,\pm 90$, or $180$ degrees around the upward-pointing z-axis.
Furthermore, we add one termination action that enables the agent to indicate that the current assembly is finished or not possible to continue, as there are no feasible actions left. Thus, the resulting action space contains $N_a = N_U \times (N_F+N_E) \times 4 + 1$ actions.
Note that the \gls{mdp} is focused on high level decision making. It does not account for the low level motion generation, namely grasp selection and robot motion planning, as this would further increase the already large action space. 
Nevertheless, given the action, the motion generation problem is well defined as it specifies the block that is to be moved, the required relative change in orientation, and its placement location.
After every placement action, all primitive units belonging to the moved block are transferred from the set of unplaced elements to the set of placed ones. We also update the set of grid-cells by removing all cells that are now occupied.

On every successful placement action, we assign a reward of $r(s_t,a_t) = 0.2 (N_{F_t}-N_{F_{t+1}} + N_{E_{t+1}}-N_{E_{t}})$, thereby giving a positive signal when the action reduced the number of target grid-cells, while also penalizing unnecessary filling of non-target grid-cells, therefore actively enforcing resource efficiency.
The conditions for a successful placement action are that the block can be placed by the robot without moving or colliding with any other block, and that it is placed in a stable configuration (i.e. the resulting structure is not falling apart due to gravity).
If the agent acting in the environment decides upon an invalid action, the episode is terminated and a reward of $-1$ is assigned.
Otherwise, the episode is terminated upon the events of i) the agent choosing the termination action, ii) no more available building blocks, or iii) the completion of \gls{rad}, i.e., the filling of all target grid-cells. As the last case corresponds to the desired behaviour, we increase the final reward by $+1$ upon this event.
Finally, to reflect the long-horizon of the considered task, we set the discount factor $\gamma$ to $0.999$.

\section{METHOD}

To reliably solve \gls{rad}, we introduce our proposed tri-level hybrid approach that efficiently handles the huge action space and combines global decision-making, considering the overall goal, with local decision-making regarding the assembly process and sequence (cf. Fig. \ref{fig:overview_alt}).
Those requirements are also inspired by the findings of our previous work \cite{funk22a} in which we discovered increasing difficulties when attempting to solve more complex \gls{rad} instances with different block types and bigger target shapes in an end-to-end fashion. 
As we attribute the difficulties to the fact that the combined learning of the global and local policy renders a challenging exploration problem, in this paper, we propose to address these issues through our novel structured hierarchical approach.
In the following, we will describe the method's all three levels, starting with the \gls{milp} formulation for resolving the global resource assignment problem, followed by a flexible, learned \gls{gnn} for task sequencing, and conclude with the low level \gls{gamp} module for handling the robotic execution on the joint level.

\subsection{\gls{milp} for optimal geometric target filling (high level)}
\label{sec:milp}

In the first step, we solve an \gls{milp} which is targeted at optimizing the blocks' placing poses to optimally fill the desired shape in light of the problem's combinatorial complexity. However, to render the problem tractable, we do not consider the sequencing and robotic constraints, therefore, only reasoning on the geometric level. 
\Gls{milp} formulations have been successfully applied for solving related tasks such as the container loading problem \cite{fasano2015modeling}, thus, in the following, we present a formulation suitable for \gls{rad}. 
Based on the grid-based parametrization of the target shape and the definition of the reward (Sec. \ref{sec:problem_def}), we can define the objective function of the \gls{milp} that is subject to maximization as
\begin{equation}
\Scale[0.85]{
    \mathcal{O}_{\text{MILP}} = \max_{\vg} \vc^T \vg,}
    \label{eq:cost}
\end{equation}
where vector $\vg$ represents the grid-state, and $\vc$ contains weighting factors that indicate whether a grid-cell should be filled or not. 
This step necessitates flattening the three-dimensional grid into a vector.
For the exemplary 2D problem displayed in Fig. \ref{fig:grid} (grid dimensions $n_x \times n_y$), the indices of all points can be flattened to a single index $j$ through $j=d_x + d_y n_x$ with the discrete coordinates of every grid-cell, $d_x$, $d_y$. Therefore, the first three entries of $\vc$ are set according to $c[0]{=}c[1]{=}-1$ and $c[2]{=}1$, as the two leftmost lowest grid-cells should not be occupied, whereas the neighboring cell on the x-axis should be.
Adapting this flattening process to the 3D case is straightforward, i.e. $j=d_x + d_y n_x + d_z n_x n_y$, with the additional grid index for the z coordinate $d_z$.
As every grid-cell can only be occupied at maximum by one primitive unit, we add the constraint
\begin{equation}
\Scale[0.85]{
    g[i] \leq 1 \quad \forall g[i] \in \vg} \text{.}
\end{equation}

Next, we need to determine how every potential positional and rotational placement action influences the grid-state.
Therefore, for each type of building block (i.e., without disambiguating between the same blocks that are just placed in a different initial pose in the environment), we first attempt placing it with all available actions and determine how the placement affects the grid-state.
For example, placing the horizontal block from Fig. \ref{fig:grid} in the lowest left position without changing the rotation results in a grid state of $\vp_{i=1,k=1}^T = [1,1,0,....,0]$, with block type index $i$ and placement action $k$. By additionally assigning an integer decision variable $w_{i,k}$ and taking all object types into account, we can define the change in the grid-state according to 
\begin{equation}
\Scale[0.85]{
    \vg = \sum_{\hat{i}=1}^{P} \sum_{\hat{k}=1}^{K(\hat{i})} w_{i=\hat{i},k=\hat{k}} \vp_{i=\hat{i},k=\hat{k}}}
\end{equation}
with a total of $P$ different block types and $K(i)$ admissible placement actions.
Note that the number of admissible placement actions is block-type-dependent, as we require that upon any placement action all primitive units stay within the grid boundaries. Considering Fig. \ref{fig:grid}, it is for instance possible to place a block consisting of a single primitive unit in the lower right-hand corner, whereas this action is not admissible for the horizontal block, as the block's right primitive unit would then end up outside the grid boundaries. 
\begin{figure*}[ht!]
    \centering
    \includegraphics[width=\textwidth]{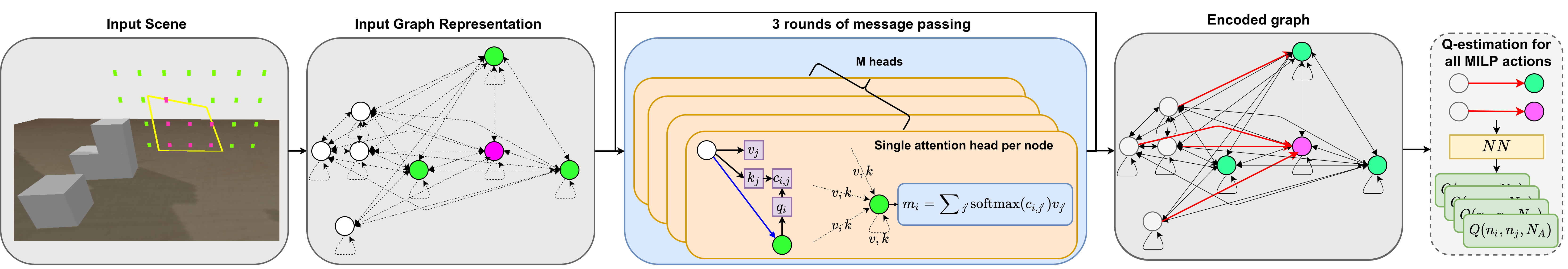}
    \vspace{-0.75cm}
    \caption{Illustrating the process of action selection using the \gls{gnn}. First, the current scene is transformed into a graph. Note that we only visualize a subset of the target (pink) and non-target (green) grid-cells. The white nodes depict the primitive units-to-be-placed. Next follow the 3 rounds of message passing in which all the nodes' features are updated using the attention mechanism (see \cite{funk22a} for details), which results in an encoded version of the graph. Finally, we perform action selection by predicting the Q-values for all the actions that are part of the \gls{milp} solution (visualized through the red arrows). The Q-values are predicted based on the nodes' features of the respective primitive units-to-be-placed and the grid-cells using a feedforward \gls{nn}.}
    \label{fig:gnn}
    \vspace{-0.65cm}
\end{figure*}

While the integer decision variables prohibit any partial block placement by definition, we still have to restrict that any block type can only be placed depending on its appearance in the current environment ($N_i$), i.e.
\begin{equation}
\Scale[0.85]{
    \sum_{\hat{k}=1}^{K(i)} w_{i,k=\hat{k}} \leq N_{i}, \quad \forall i \in P}.
\end{equation}
This constraint concludes the \gls{milp} formulation (optimize (1), with constraints (2-4)), which boils down to optimizing the integer decision variables $w_{i,k}$ 
through Gurobi \cite{gurobi}.
Thus, the solution contains the quantity and final poses for every block type and is guaranteed to optimally fill the desired target shape. Yet, it neither resolves the challenge of determining the assembly order nor the combinatorial ambiguity regarding which block to use for each placement, as the scenes typically contain multiple blocks of same type.

\subsection{GNN for task sequencing (medium level)}

The high level \gls{milp} only partially resolves the combinatorial aspect of \gls{rad}. It lacks the placement actions' sequencing and the exact assignment of which block to use for each placement. 
Further, the \gls{milp} does not consider robotic feasibility, the blocks' initial positions, and neither structural stability during assembly.
We thus require another level, capable of efficiently incorporating the \gls{milp}'s prior knowledge, and deciding upon either executing one of the proposed actions or terminating the current assembly if none of the proposed actions is feasible. 
This can either be due to robotic constraints or due to stability considerations (e.g., robot colliding with the structure while placing a block / creating an unstable structure).
Still, having the \gls{milp}'s solution allows for efficiently shrinking the action space that has to be considered on this level and thus allows for using an approach based on \textit{guided exploration} and model-free \gls{rl}, capable of reflecting all real-world constraints without requiring any further simplifications. 

Thus, we propose to employ an approach based on a combination of \gls{gnn} and Q-learning \cite{hamrick2019combining, funk22a}, for the following reasons.
The graph-based representation is capable of providing the required flexibility on the representation level and invariance to the problem size, while performing the action selection based on Q-learning is desirable as i) the herein considered action space is discrete, but remains tractable for exploration due to the prior knowledge from the \gls{milp} solution, ii) the state-action-based formulation allows to efficiently incorporate the prior knowledge by masking out all actions that are not inside the \gls{milp} solution, iii) potential multimodalities in the \gls{milp} solution are not problematic and do not erroneously bias this Q-function estimator, and iv) it allows easy and time-effective combination with search-based methods, such as \gls{mcts} to improve robustness and performance despite the combinatorial action-space \cite{silver2017mastering}. 
Moreover, since the overall method only requires running the potentially time-consuming process (depending on problem size) of solving the \gls{milp} once, this level should be reactive w.r.t. changes in the blocks' positions.

We now briefly describe the process of action selection, as also visualized in Fig. \ref{fig:gnn}, but refer to our previous work \cite{funk22a} for the additional details. 
We first transform the environment's current state into a graph, by creating nodes for all primitive units and grid-cells (cf. Sec. \ref{sec:problem_def}). 
The nodes' features 
contain the respective nodes' 3D position, as well as 2 indices that indicate the node type, i.e. placed/unplaced primitive unit, target/non-target grid-cell.
Almost all nodes of the graph are fully-connected with each other -- we only omit the connections in-between the unplaced primitive units if they do not belong to the same block for the purpose of explicitly encoding different blocks. 
Note that the edges solely define in between which nodes there is exchange of information.
Upon graph creation follow three rounds of message passing using an attention mechanism \cite{funk22a, kool2018attention}.
This process updates the nodes' values based on their neighbors' and can also be seen as building a meaningful graph encoding due to the flow of information.
The obtained, updated node values are then used as the basis for computing Q-values for all available actions. 
Particularly, as we can place any unplaced primitive unit w.r.t. every grid-cell, for this final step, we use a standard feedforward \gls{nn} that takes as input the encoded node values of i) the primitive unit-to-be-placed, and ii) the grid-cell, and outputs the Q-values for all the four rotational-placement actions in between these nodes. 
That way, we predict the quality of moving the primitive unit to the grid-cell, including the potential re-orientation. Note that this action moves the entire block that the primitive-unit-to-be-placed is part of. 
This process is repeated for all pairs of unplaced primitive units and grid-cells. To compute the Q-value for the termination action, we feed the average of all nodes' features through a different \gls{nn}.

Action selection is done using an $\epsilon$-greedy strategy, yet, only allowing to choose actions from the \gls{milp} solution and the termination action.
Originally, the \gls{milp}'s result only contains the final poses and quantities per block type (cf. Sec. \ref{sec:milp}).
To make this information compatible with the current level, we infer all the translational and rotational actions w.r.t. the primitive units that will put the respective block into all the desired final poses.
This ensures that all blocks of each type are considered for every placement, %
and it also directly yields the allowed actions between primitive units and grid-cells (cf. Fig. \ref{fig:gnn}).
After every placement, all actions that would put another block in the same pose are removed.
Note that the \gls{milp}'s prior knowledge is not considered at an earlier stage, as we view the message passing process as creating an holistic understanding of the \gls{rad} scene. Thus, we only exploit it on action selection.

The graph's weights are refined through temporal-difference learning. In particular, we minimize the smooth L1 loss between the current Q-value prediction of the \gls{gnn} and the estimated value based on collected rollouts and a target network $Q_T$, i.e. $\hat{Q}(s_t,a_t)=r(s_t,a_t)+\gamma \max_{\Tilde{a}} Q_T(s_{t+1},\Tilde{a})$, where the reward is defined according to Sec. \ref{sec:problem_def}.
While this Q-learning procedure by itself already results in good policies that can directly be used for action selection, at test time, we can additionally consider action selection based on the combination of Q-learning and \gls{mcts} (DQN+MCTS).
This combination has proven very effective when dealing with combinatorial action spaces, and, in particular, improves policy performance, transferability and robustness \cite{silver2017mastering, hamrick2019combining}.
The combination is especially attractive, as the Q-function allows bootstrapping the depth of the Monte Carlo rollouts, which in turn allows exploring the effect of different actions without requiring costly simulated deep rollouts until episode termination. Again, following \cite{funk22a}, we start the search from the current state by choosing an action according to an $\epsilon$-greedy strategy and terminate every rollout directly after the first action, and estimate the next state's quality using the Q-function.

To conclude, in this intermediate level, we make use of a policy based on the combination of \gls{gnn}, \gls{rl}, and eventually add \gls{mcts} during test time. The policy benefits from the reduced combinatorial action space, determines the sequencing, and assigns the blocks to the  placements, while considering all the environmental constraints.
The last part that is missing, is a policy that turns these commands into joint-level signals to actually move the robot and the blocks.

\subsection{Robot grasp and motion planning (low level)}

The lowest level is tasked with converting the previous level's actions into robot joint commands, thereby handling the final robot execution of block grasping and placing.
While it would be possible to add those decisions to the previous level, we consider motion generation separately, as it heavily depends on the actual robot manipulator, and we do not want to further increase the previous level's action space.
Due to the scenes' layout (Fig. \ref{fig:overview_alt}), i.e., the blocks being close together initially and during the final placement, we first check the feasibility of a predefined set of top-down grasping poses and subsequently check if this grasp results in a feasible final placement pose. 
An action is feasible if the requested grasping/placement pose can be reached by the robot, i.e., considering the joint limits, and that there are no collisions with the other blocks in the scene in this configuration (which is computed using \gls{ik}).
If there exists a pair of feasible grasping and placing poses, we move the robot by approaching the grasping pose from the top, then move to a position that is slightly above the placing location, and finally, approach the placement pose. Again, all intermediate waypoints are computed based on \gls{ik}.

\section{EXPERIMENTAL RESULTS}

\begin{figure*}
    \centering
    \includegraphics[width=\textwidth]{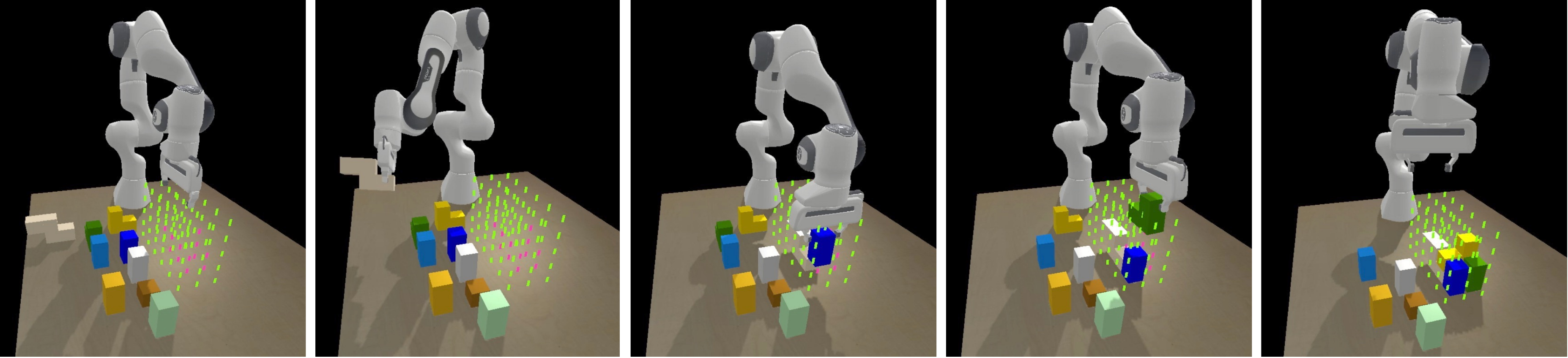}
    \vspace{-0.75cm}
    \caption{\small Illustration of a successful \gls{rad} sequence using our proposed MILP-DQN-MCTS approach. The agent successfully the assembly successfully using in total 4 blocks and 3 different block types.}
    \label{fig:ass_mcts}
    \vspace{+0.15em}
    \centering
    \includegraphics[width=\textwidth]{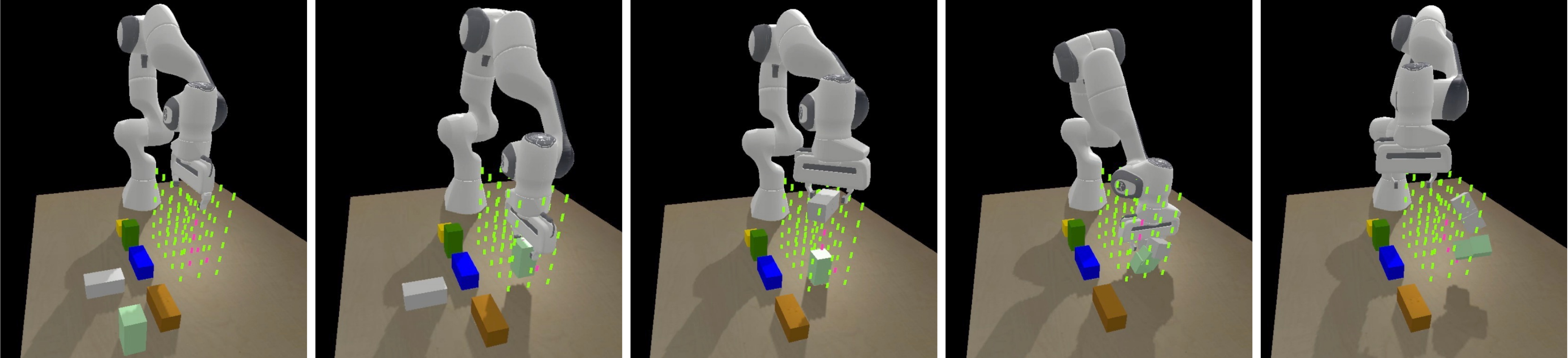}
    \vspace{-0.75cm}
    \caption{\small Illustration of an unsuccessful \gls{rad} sequence using the heuristic agent (HEUR) introduced in Sec. V-B. As shown in the images, it is important to perform informed decisions about the assembly sequence, as the wrong sequencing can result in collisions between the block that is placed and other blocks in the scene.}
    \label{fig:ass_heur}
    \vspace{-0.65cm}
\end{figure*}

We now present the experimental evaluation of our proposed \gls{milp}-DQN method and potentially adding \gls{mcts} with a search budget of 5 (\gls{milp}-DQN-\gls{mcts}), as in \cite{funk22a}. 
We first evaluate in simulated RAD environments (using PyBullet \cite{coumans2021}, cf. Figs. \ref{fig:ass_mcts} \& \ref{fig:ass_heur}) for answering two questions: 
1) Does the \gls{milp}'s guiding exploitation signal help to effectively boost the proposed method's performance compared to employing an end-to-end approach without any prior, i.e., is the high level \gls{milp} really needed?
2) How effective is the \gls{gnn} policy of the medium level compared to using a heuristic approach for task sequencing, i.e., is the medium level \gls{gnn} required?
Lastly, we investigate whether our policies can be transferred to the real world.

Before diving into the results, we quickly explain the training procedure.
As the training of our proposed approach requires knowledge of the \gls{milp} solution for every RAD scene, we decided to create a dataset prior to training. 
This dataset contains $50,000$ different scenes, describing the environment's initial state, upon which the agent is subsequently acting, modifying its state through the actions.
For the experiments that do not consider the robot, we train all agents for $1,000,000$ steps, while we train for $1,500,000$ steps in the robot experiments. For all methods we train 5 agents using different seeds \cite{d2021mushroomrl}.
We describe the difficulty of our two-sided \gls{rad} environments (i.e., two target shapes must be filled) by specifying the maximum height and width of the admissible target shapes, e.g., Fig. \ref{fig:grid} shows a potential one-sided target shape of height and width 3.

We will use a star(*) to denote the agents' evaluation in their training conditions, i.e., using target shapes of similar size, yet, using different initial environment states (i.e. blocks and poses) compared to training. The other experiments are even further out-of-distribution, as the target shapes are guaranteed to be bigger compared to the ones seen during training. As bigger shapes require more blocks, the number of initially placed blocks is also increasing. 
The results are obtained by averaging the agents' performance in 200 scenes.
We report the discounted reward $R$, the fraction of runs that ended i) upon perfectly recreating the target shape $d$, ii) with selecting the termination action 
$e$, iii) upon failure $f$, i.e., trying to execute an action that is not feasible with the robot, or placing the block in an unstable configuration, or destroying the already existing structure, while differentiating between failing on grasp selection $f_g$, i.e., no feasible grasp exists, and on block placement $f_p$ for the robot experiments. Finally, we report the desired target grid-cell coverage $\bar{a}$, i.e., the fraction of target grid-cells that were initially supposed to be filled and have actually been filled. 
We also provide videos of the experiments at {\small \url{https://sites.google.com/view/rl-meets-milp}}.

\noindent\textit{A) \underline{Is the high level \gls{milp} needed?}}
\\
We consider a scenario without the robot-in-the-loop, which reduces the complexity as \gls{gamp} can be omitted. 
Thus, the task reduces to placing the blocks in a stable configuration while trying to optimally fill the desired shape.
We compare our approach against two baselines that do not consider the \gls{milp}.
The first one (DQN) can place any of the available blocks at all currently unoccupied grid-cells.
The second one (DQN-REL) follows our previous work \cite{funk22a}, in which the blocks can only be placed next to the already placed ones, thereby reducing the action space.
In the first step only, we allow placing the blocks at any target grid-cell.
\\
\begin{table}[t]
\begin{center}
\caption{\small Comparing our proposed method with two learned baselines in the two-sided environment wo robot.} 
\label{table:wo_robo}
\scriptsize
\scalebox{0.9}{
\begin{tabular}{l|l|cccc}
    Grid Size & Method  & $R$ &  $e$ & $d$ & $\bar{a}$  \\
\hline
\hline
{3*} & 
DQN                                               &0.63 (0.02) & 0.59 & 0.27 & 0.71     \\                                            
                     &DQN-REL \cite{funk22a}   
                     &0.67 (0.01) & 0.7 &  0.23 & 0.68
                          \\
                     &\textbf{MILP-DQN}   
                     &\textbf{1.22} (0.01) & 0.31 & \textbf{0.53} & \textbf{0.87}
                            \\

\hline

{4} & 
DQN                                               &0.71 (0.08) & 0.53 & 0.2 & 0.69     \\                                            
                     &DQN-REL \cite{funk22a}  
                     &0.75 (0.08) & 0.65 &  0.14 & 0.66
                          \\
                     &\textbf{MILP-DQN}  
                     & \textbf{1.56} (0.03) & 0.25 & \textbf{0.47} & \textbf{0.87}
                            \\
\hline
{5}                     &\textbf{MILP-DQN}  
                     & \textbf{1.92} (0.05) & 0.17 & 0.42 & \textbf{0.85}
                            \\
\hline

\end{tabular}}
\end{center}
\vspace{-0.95cm}
\end{table}
The results in Table \ref{table:wo_robo} reveal that the \gls{milp} provides a strong inductive bias that is effective in guiding the exploration, as the agents trained using our proposed MILP-DQN approach outperform the two baselines.
The baseline agents exhibit very similar performance, with DQN-REL yielding slightly higher rewards.
Actually, the smaller action space of the DQN-REL agents results in better performance at the beginning of the training.
Compared to the baseline agents, the agents trained using MILP-DQN achieve an increase in the success rate and discounted reward by a factor of 2 (grid size of 3, 4).
These results confirm the task's combinatorial complexity. Performing an $\epsilon$-greedy exploration without using an informed prior does not allow for discovering good action sequences. Therefore, the baseline agents learn to terminate more frequently and achieve significantly lower successes and rewards.
The results also reveal that the MILP-DQN agents generalize well to the out-of-distribution environments as the desired target grid-cell coverage remains high at 0.87 and 0.85 (grid size of 4,5), despite the significant increase in task complexity, i.e., the average target grid-cells that should be filled increase from roughly 5 to 12 while increasing the grid size from 3 to 5. 
We also want to point out that some desired shapes contain configurations of target grid-cells and non-target grid-cells, such that the \gls{milp}'s solution does not contain all the actions that would be needed to achieve $\bar{a}{=}1.0$ as the \gls{milp} optimizes a tradeoff between optimal coverage and resource efficiency.
When correcting for this effect in the computation of $\bar{a}$, the values increase to 0.97, 0.91 \& 0.87 for MILP-DQN (grid sizes of 3, 4 \& 5).

\noindent\textit{B) \underline{How effective is the \gls{gnn} policy for robotic execution?}}
\label{sec:exp_w_robot}
We now consider scenarios with the robot-in-the-loop (Figs. \ref{fig:ass_mcts}, \ref{fig:ass_heur}) and investigate the effectiveness of the medium level \gls{gnn} policy.
For this purpose, we compare the learned \gls{gnn} with a heuristic (HEUR).
The agents using HEUR perform action selection as follows: based on actions proposed by the \gls{milp}, the heuristic only considers those for which the block's placement will result in a stable configuration, i.e., all grid-cells below the block-to-be-placed are already filled. Subsequently, the HEUR selects one action from this subset at random. If there is no action available that satisfies these conditions, the termination action is selected.
Additionally, we report the performance of the heuristic agent on the same problem instances, however, without using the robot for part placement (HEUR wo robot). We consider this agent as an oracle, as it is acting in a substantially simpler environment without having to consider any robotic constraints.

The results from these experiments are presented in Table \ref{table:w_robo}.
In both versions of the environment, there is a significant difference between the oracle heuristic (HEUR wo robo) and the heuristic baseline (HEUR) that underlines the increased difficulty from having the robot-in-the-loop. Moreover, our proposed MILP-DQN \& MILP-DQN-MCTS agents outperform the heuristic baseline (HEUR).
Notably, already in the environment with less building blocks, i.e. with the grid size of 4, using the heuristic on the medium level results in 40\% of all the rollouts terminating upon an invalid action.
Slightly more failures, i.e., 24\% can be attributed to grasp selection, i.e. the policy selecting a block for placement that cannot be grasped or placed without collisions, while 16\% are due to collisions during placement. 
Such a failure is depicted in Fig. \ref{fig:ass_heur}, where due to bad action sequencing by the HEUR agent, the two blocks collide.
Those high rates of failure indicate that a more informed method for action sequencing is actually required. 
As can be seen in Table \ref{table:w_robo}, both versions of our proposed approach are capable of effectively reducing the percentage of failures, with MILP-DQN decreasing the rates roughly by a factors of 3 \& 2 (for grasping and placing failures, respectively), while the addition of \gls{mcts} leads to an impressive decrease by factors of 12 \& 5. 
\begin{table}[t]
\begin{center}
\caption{\small Comparing our proposed method with a heuristic in the two-sided environment with the robot-in-the-loop.} 
\label{table:w_robo}
\scriptsize
\scalebox{0.95}{
\begin{tabular}{l|l|cccccc}
    \multicolumn{1}{l|}{Grid} & & & & & & \\
    Size & Method  & $R$ &  $e$ & $f_g$ & $f_p$ & $\bar{a}$  \\
\hline
\hline
{4*} & HEUR (wo robo)                                               &- & - & - & - & 0.81     \\  
&HEUR                                               &0.57 (0.04) & 0.36 & 0.24 & 0.16 & 0.62     \\  
                     &\textbf{MILP-DQN}   
                     &1.03 (0.04) & 0.49 &  0.08 & 0.08 & 0.7
                          \\
                     &\textbf{MILP-DQN-MCTS}   
                     &\textbf{1.24} (0.03) & 0.57 & \textbf{0.02} & \textbf{0.03} & \textbf{0.75}
                            \\

\hline

{5} & HEUR (wo robo)                                               &- & - & - & - & 0.78     \\  
&HEUR                                               &0.34 (0.02) & 0.29 & 0.34 & 0.24 & 0.47     \\
                     &\textbf{MILP-DQN}   
                     &0.98 (0.06) & 0.53 &  0.1 & 0.15 & 0.58
                          \\
                     &\textbf{MILP-DQN-MCTS}   
                     &\textbf{1.38} (0.04) & 0.65 & \textbf{0.02} & \textbf{0.06} & \textbf{0.65}
                            \\
\hline

\end{tabular}}
\end{center}
\vspace{-0.95cm}
\end{table}
Those results show that our learned graph-based representations are indeed capable of effectively capturing the state of the environment and make informed decisions regarding the action sequencing which is a crucial component of \gls{rad}. 
The clear advantages also prevail for the even more difficult scenarios, considering the grid sizes of 5, where again, both versions of our proposed algorithm also achieve significantly higher rewards and target grid-cell coverage compared to the HEUR baseline.
While there is a slight drop in performance concerning the achieved coverage of the MILP-DQN-MCTS agent with the increase in environment difficulty, the increase in failures is marginal (only by 3\% on placing).
Moreover, when relating the target grid-cell coverage of MILP-DQN-MCTS with the performance of the oracle HEUR wo robot agent, our proposed agents achieve relative fillings of 0.93 and 0.83 respectively, therefore again underlining their effectiveness.
For more complicated scenes with grid sizes of 6, the performance of MILP-DQN-MCTS slightly degenerates. Yet, we attribute this behavior to the combination of extremely cluttered scenes and the robot's limited workspace, and speculate that mobile manipulators could circumvent these issues given the strong generalization from the previous experiments.
Overall, the experiments show that our proposed hierarchical approach is indeed capable of resolving the inherent difficulties of \gls{rad}, as also illustrated in Fig. \ref{fig:ass_mcts} where we show the successful assembly of a desired target shape using 4 blocks of 3 different types.

\noindent\textit{C) \underline{Is real-world policy transfer possible?}}
\label{sec:hw_exp}
\begin{figure}
    \centering
    \includegraphics[width=1.0\columnwidth]{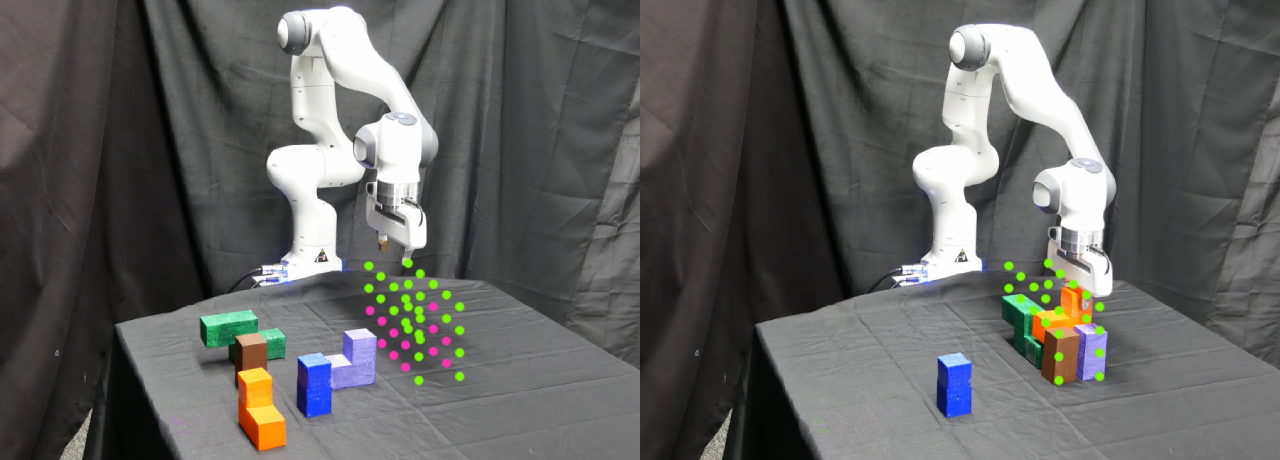}
    \vspace{-0.75cm}
    \caption{\small Real-world \gls{rad}. Given the initial configuration (left), our proposed MILP-DQN-MCTS yields a valid assembly sequence that ends up filling all the desired target grid-cells (right).}
    \label{fig:hw_experiments}
    \vspace{-0.5cm}
\end{figure}

\noindent Finally, we evaluate whether the obtained MILP-DQN-MCTS policies can be transferred to real-world \gls{rad} scenes (cf. Fig. \ref{fig:hw_experiments}).
For the evaluation, we first register all of the building blocks' poses using OptiTrack and initialize a simulated \gls{rad} scene mirroring the real-world.
The simulated twin environment is subsequently exploited for evaluating our policies and performing \gls{mcts} planning to decide upon the next action which is then executed in both, simulation and reality.
As shown in Fig. \ref{fig:hw_experiments} and in the supplementary videos, we find that our proposed MILP-DQN-MCTS agents can indeed be transferred to real-world \gls{rad} scenes. This once again underlines its robustness w.r.t. different scenes, in particular, w.r.t. scene initializations and part placements.

\section{CONCLUSION}

We have presented a novel hierarchical approach for robot assembly discovery (RAD).
Our proposed approach is based on the powerful combination of global reasoning through mixed-integer programming, with graph-based reinforcement learning and model-based search for local decision-making, together with grasp and motion planning for realizing the assembly actions on the manipulator's joint level.
The hierarchy allows for the efficient decomposition of the original problem's huge combinatorial action space and thereby results in robust, reliable, and effective \gls{rad} policies. 
The proposed approach is validated in a set of simulated \gls{rad} experiments and achieves an average coverage of the desired target shape of 75\% while maintaining extremely low rates of failure (5\%). We also showcase transfer to real-world \gls{rad} scenes.
In the future, we want to investigate how this approach can be scaled to handle a wider range of objects.

\bibliographystyle{IEEEtran}
\bibliography{references.bib}

\end{document}

%% file: misc/math_commands.tex
\def\1{\bm{1}}

\def\vc{{\bm{c}}}

\def\vg{{\bm{g}}}

\def\vp{{\bm{p}}}

\def\vx{{\bm{x}}}

\DeclareMathAlphabet{\mathsfit}{\encodingdefault}{\sfdefault}{m}{sl}
\SetMathAlphabet{\mathsfit}{bold}{\encodingdefault}{\sfdefault}{bx}{n}

\makeatletter
\newcommand{\StatexIndent}[1][3]{%
  \setlength\@tempdima{\algorithmicindent}%
  \Statex\hskip\dimexpr#1\@tempdima\relax}
\makeatother